# Automated Generation and Tagging of Knowledge Components from Multiple-Choice Questions

Steven Moore
Human-Computer Interaction
Carnegie Mellon University
Pittsburgh, PA, USA
StevenJamesMoore@gmail.com

Robin Schmucker
Machine Learning Department
Carnegie Mellon University
Pittsburgh, PA, USA
rschmuck@cs.cmu.edu

Tom Mitchell
Machine Learning Department
Carnegie Mellon University
Pittsburgh, PA, USA
tom.mitchell@cs.cmu.edu

John Stamper
Human-Computer Interaction
Carnegie Mellon University
Pittsburgh, PA, USA
jstamper@cmu.edu

## ABSTRACT

Knowledge Components (KCs) linked to assessments enhance the measurement of student learning, enrich analytics, and facilitate adaptivity. However, generating and linking KCs to assessment items requires significant effort and domain-specific knowledge. To streamline this process for higher-education courses, we employed GPT-4 to generate KCs for multiple-choice questions (MCQs) in Chemistry and E-Learning. We analyzed discrepancies between the KCs generated by the Large Language Model (LLM) and those made by humans through evaluation from three domain experts in each subject area. This evaluation aimed to determine whether, in instances of non-matching KCs, evaluators showed a preference for the LLM-generated KCs over their human-created counterparts. We also developed an ontology induction algorithm to cluster questions that assess similar KCs based on their content. Our most effective LLM strategy accurately matched KCs for 56% of Chemistry and 35% of E-Learning MCQs, with even higher success when considering the top five KC suggestions. Human evaluators favored LLM-generated KCs, choosing them over human-assigned ones approximately two-thirds of the time, a preference that was statistically significant across both domains. Our clustering algorithm successfully grouped questions by their underlying KCs without needing explicit labels or contextual information. This research advances the automation of KC generation and classification for assessment items, alleviating the need for student data or predefined KC labels.

## 1 INTRODUCTION

Digital learning platforms, such as MOOCs and interactive online courses, facilitate the mapping of assessments to specific skills or competencies, referred to as Knowledge Components (KCs). These KCs are instrumental in driving learning analytics systems, enabling adaptive content sequencing, and providing precise estimates of student mastery levels [7,9,16]. KCs embody the cognitive functions or structures inferred from student performance on related assessments and are more nuanced and fine-grained than broad course learning objectives [17]. The process of associating assessments with KCs, often referred to as skill or concept tagging, generates a comprehensive map of the knowledge conveyed by the platform or course. This mapping, termed Knowledge Component Model (KCM), is crucial for closely monitoring student learning, allowing educators to identify precisely which concepts a student may find challenging based on their assessment performance [39]. Without an accurate KCM, the effectiveness of assessing mastery and implementing adaptive learning strategies may be significantly hindered [35].

While KCMs offer numerous benefits for digital learning platforms, the creation of KCs for each assessment is a time-intensive process that demands domain expertise. This usually requires a domain expert, such as the course instructor, to determine the necessary KCs for solving each assessment [11]. This process necessitates the identification of at least one KC per assessment item and can involve identifying up to three or more KCs, depending on the complexity of the assessment, the subject area, and the educational level [8]. Modifying existing KC tags is equally challenging; although evaluating the effectiveness of KCMs is feasible, adjusting it can be as time-consuming as the initial creation process. Evaluation methods such as learning curve analysis and statistical measures such as Akaike Information Criterion (AIC) can provide insights into the effectiveness of the KCMs [18]. However, these methods often require large amounts of data to produce reliable results, and primarily focus on the fit and predictive accuracy of the model, potentially overlooking the contextual and practical applicability of the KCs in diverse educational environments. The continual evaluation and updating of these models remain critical, as many learning platforms in the United States have begun transitioning to common taxonomies, like the Common Core State Standards, necessitating the re-tagging of content or the alignment of existing mappings with these standards [40]. This re-mapping is not only labor-intensive but also needs to be revisited with each update or change to the common standards, adding to the ongoing workload.

To mitigate the challenges associated with mapping KCs to assessments, a variety of automated solutions employing machine learning (ML) and natural language processing (NLP) have been proposed [13,34]. These approaches primarily employ classification algorithms, using an existing repository of KCs as reference labels for mapping. However, one critical limitation of these techniques is their reliance on a predefined set of KCs. They are not designed to identify new KCs, but instead depend on a pool of KCs, which may not always be available. The difficulty of automatically generating new KCs lies in ensuring their specificity and relevance to the problem at hand, their relationship with other KCs, and their alignment with the overall course content [47]. Consequently, while the challenge of associating assessments with existing KCs is significant, the task is further complicated by the initial requirement to generate these KCs, a step that previous efforts often overlook. When previous studies involve KC generation, they are frequently unlabeled, necessitating a human to create their descriptive text labels [5,9].

Recent advancements in large language models (LLMs) have shown promise in automating the generation and tagging of metadata to educational content [41]. To explore this possibility in the context of KCs, we utilize datasets from two different domains: Chemistry and E-Learning, encompassing the higher-education levels of undergraduate and masters. The respective datasets contain multiple-choice questions (MCQs), with each question mapped to a single KC. We leverage GPT-4 [31], a state-of-the-art LLM, to generate a KC for each MCQ by employing two distinct prompting strategies, based solely on the text of the MCQs. We then compare LLM-generated KCs to the original human-assigned ones, which serve as our gold standard labels. For KCs that did not match, we had groups of three human domain experts evaluate the discrepancies to determine their preferred KCs, which often leaned towards the ones generated by the LLM. Additionally, to organize the KCs, we implement a clustering algorithm that leverages the content of the MCQs to group questions that assess the same KCs. Our research presents a methodology that automates the process of generating and tagging KCs to problems across various domains, relying exclusively on the text of the assessments.

The main contributions of this work are: 1) A proposed method for generating KCs for assessments using LLMs, 2) Empirical and human validation of the LLM-generated KCs, and 3) A technique that iteratively induces a KC ontology and that clusters assessments accordingly.

## 2 RELATED WORK
### 2.1 Knowledge Component Models

A Knowledge Component Model (KCM) acts as a foundational framework in education, particularly in digital learning, to systematically organize and define the KCs students are expected to acquire [39]. KCs are generally characterized by their detailed and fine-grained nature, yet they can be modeled at various levels of granularity. The Knowledge-Learning-Instruction (KLI) Framework provides flexibility in determining the specificity with which KCs are defined [17]. However, the primary objective of a KCM is to elucidate the relationships among these KCs and how they connect to assessments or learning activities in an educational program or course. A KCM enables the monitoring of student progress, the prediction of learning outcomes, and the identification of effective teaching strategies through learning analytics [15]. Additionally, KCMs underpin adaptive learning systems, which tailor the content, difficulty level, and pacing according to the learner's mastery of the KCs, thereby seeking to enhance learning efficiency and outcomes [14].

Validation of successful KCMs traditionally relies on a mix of automated and manual metrics such as Learning Curve Analysis (LCA) and Item Response Theory (IRT), or evaluations by human domain experts [1]. The former automated methods are robust and objective indicators of a KCM's validity, but they require student performance data on the assessments. For example, LCA visualizes students' performance over time or across learning opportunities, where a well-validated KCM shows consistent improvement as students engage with KC-aligned content [26]. Discrepancies, such as sudden leaps or plateaus, may suggest the need for model refinement. IRT models further validate KCMs by examining the relationship between students' latent traits, like mastery of KCs, and their performance on specific assessments, expecting a strong correlation between KC mastery and correct responses [46]. On the other hand, human evaluation involving domain experts' review of the KCM is a commonly used qualitative method that involves more subjectivity but does not require student performance data [24]. Properly validated KCMs enhance adaptive learning systems and analytical tools, allowing for more accurate measurement of student mastery, while also offering valuable insights into potential areas for improvement within the instructional design process [22].

## 2.2 Human KCM Generation

Traditional approaches to generating KCMs involve manual processes, rooted in either theoretical or empirical task analysis [25]. Theoretical methods require the domain expert to meticulously examine a range of activities and instructional materials to identify the knowledge requirements of the task [7]. This process can be supported by tools (i.e. CTAT [48]) or frameworks (i.e. EAKT [37]), that can facilitate the identification of different problem-solving steps and the associated KCs that learners must possess. Although technology can aid in this process, it is not strictly necessary; the task can also be accomplished using simple methods like pencil and paper to note down the KCs hypothesized to be assessed by an activity. On the empirical side, authors gather data on problem-solving within a specific task domain through methods like contextual inquiry, think-aloud protocols, and cognitive task analysis (CTA) [3]. In particular, CTA typically involves instructional designers or learning engineers prompting domain experts to explain their thought processes while engaging in a series of tasks. This approach helps in uncovering the underlying KCs that are essential for task performance.

Human-generated KCMs are pivotal in the deployment of various educational technologies, yet their creation presents multiple challenges. The process, whether theoretical or empirical, is time-consuming, demands domain expertise, and often necessitates collaboration between multiple individuals, thereby introducing additional logistical complexities [1]. Even commonly used methods, such as CTA, are susceptible to the issues of human subjectivity and expert blind spots [29]. These issues can lead to potential oversights or overly generalized interpretations of KCs, resulting in omissions or inaccurately defined KCs that fail to address common novice pitfalls. Moreover, the reliance on specialized human knowledge significantly complicates the scaling of these methods. Recent endeavors have explored crowdsourcing to harness scalable human judgment for tagging Math and English writing problems with KCs [28]. However, their findings indicate that crowdworkers often struggle to achieve the level of specificity required for accurately defining KCs, highlighting a critical gap between the need for detailed knowledge articulation and the capabilities of human contributions.

## 2.3 Automated KCM Generation

The growing limitations of manual KCM construction, which relies exclusively on human input, underscore the need for more effective approaches, as manual methods often demand significant resources and time. Automated approaches for generating KCMs have emerged as valuable tools that can enhance, rather than replace, human efforts [4]. These methods employ data-driven approaches, like Learning Factors Analysis (LFA) and Q-matrix inspection, to categorize questions under existing KCs within a predefined search space in educational software [5,9,44]. They do not create new KCs, but rather assign questions to the already defined ones. Notably, models developed or improved through automated or semi-automated techniques frequently surpass their manually constructed counterparts, especially in predicting student performance [1]. For example, evidence from previous research shows that a KCM refined through a combination of human judgment and automated methods can enable students to achieve mastery 26% faster [18].

Automated approaches for generating KCMs that operate largely without human intervention typically follow two main strategies: generation or classification [30]. In terms of generation, significant efforts focus on creating knowledge graphs or extracting concepts from digital textbooks, in addition to deriving KCs from student performance data [10], for example via matrix factorization [6,12] and VAE-based methods [32]. However, these methods often face challenges related to interpretability, not only due to the opaque nature of the algorithms used, but also because the generated labels may not hold meaningful insights for educators [43]. On the classification front, the goal is to assign existing KCs to problems based on semantic information contained in the assessment text, a process that has proven effective in domains like Math and Science [36,45]. However, for areas without a well-defined standard or an established bank of KCs, such as those outside the common core standards, this classification approach presents a significant challenge due to the absence of predefined labels for categorization [20]. Another related problem is establishing the equivalence of individual KCs across different learning platforms which often use varying nomenclature to refer to the same learning objectives. Prior work explored the application of machine translation techniques that consider assessment context and textual content to identify equivalent KC pairings [21].

## 3 METHODS

### 3.1 Datasets

In this study, we utilized two datasets from higher education MOOCs: one in Chemistry and the other in E-Learning. Each dataset comprises 80 multiple-choice questions (MCQs), with each question offering between two to four answer options. These datasets are structured such that each KC is represented by exactly two MCQs, totaling 40 KCs in each dataset. The ground truth KCs associated with each question were previously identified by domain experts who contributed to developing the content and authoring the courses. A key selection criterion for these questions was that each should be associated with a single KC, ensuring clarity in mapping.

The Chemistry dataset[1] originates from an online course adopted by various universities across the United States as instructional materials for an undergraduate introductory Chemistry course. This content is hosted on a widely used digital learning platform and is often integrated into a flipped-classroom model, supplementing in-person instruction. Similarly, the E-

---

[1] https://pslcdatashop.web.cmu.edu/DatasetInfo?datasetId=4640

Learning dataset[2] is derived from a master's level course at a university in the eastern United States, utilizing the same digital platform for implementation. Both datasets and the code utilized in this study are available for inspection[3].

## 3.2 Prompting Strategies

To automatically generate KCs, our research employed the `gpt-4-0125-preview`[4] API, chosen for its speed, reduced cost, and the consistency it offers. This decision was made to avoid the variability that might arise from continuous updates to the standard GPT-4 API, which could impact the reproducibility of our results. Following recommendations from existing literature [2,38], we developed two prompting strategies to generate the KCs for each question: the *simulated expert* approach and the *simulated textbook* approach. For both strategies, the only contextual information provided to the language model was the course's educational level (undergraduate or master's) and the subject area (Chemistry or E-Learning). For each approach we supplied a MCQ, which included the question text, the correct answer, and the alternative options. This choice was motivated by our aim to develop a method that could be generalized, recognizing that providing extensive contextual information might not always be feasible for certain question banks or assessment content. This is also aligned with previous research that has attempted to generate KCs based solely on the content's text using automated NLP-based approaches [27,42]. The exact prompts used for the *simulated expert* strategy are illustrated in Figure 1 and the prompts for the *simulated textbook* strategy are in Figure 2.

```
"""Simulate three experts collaboratively evaluating a
college level multiple-choice question to determine
what knowledge components and skills it assesses. The
three experts are brilliant, logical, detail-oriented,
nit-picky {subject} teachers. The multiple-choice
question is used as a low-stakes assessment as part of
an {context} {subject} course that covers similar
content. Each person verbosely explains their thought
process in real-time, considering the prior
explanations of others and openly acknowledging
mistakes. At each step, whenever possible, each expert
refines and builds upon the thoughts of others,
acknowledging their contributions. They continue until
there is a definitive list of five knowledge and
skills required to solve the question, keeping in mind
that the question is for a college audience with
existing prior knowledge. Once all of the experts are
done reasoning, share an agreed conclusion.

Question text: {question_text}
Correct answer: {answer_text}"""

"""Based on the reasoning from these three experts and
their conclusion, reword these five points to begin
with action words from Bloom's Revised Taxonomy.
Reasonings: {reasonings}"""
```

```
"""Reasonings: {reasonings}
Five points: {points}

Of these five points, which one is the most relevant
to the question?"""
```

**Figure 1: Three prompts used for the *simulated expert* prompting strategy for KC generation.**

```
"""Below there is a multiple-choice question intended
for a {context} audience with existing prior knowledge
on the subject of {subject}. The question is used as a
low-stakes assessment as part of an {context}
{subject} course that covers similar content. The
first answer choice, option A), is the correct answer.
If this question was presented in a textbook for an
{context} {subject} course, what five domain-specific
low-level detailed topics would the page cover? Note
that the question is for a college audience with
existing prior knowledge in {subject}.

Question text: {question_text}
{options_text}"""
```

```
"Based on these topics, reword them to begin with
action words from Bloom's Revised Taxonomy, while
keeping them domain-specific, low-level, and
detailed."
```

```
"Of these topics, which is the most relevant to the
question?"
```

**Figure 2: Three prompts used for the *simulated textbook* prompting strategy for KC generation.**

In the first prompting strategy following the *simulated expert* approach, we employed the tree-of-thought technique, directing the LLM to emulate a discussion among three expert instructors [23]. The objective was to identify five specific, detailed skills and knowledge assessed by the provided MCQ. After this simulated discussion, the experts were expected to produce a list of the key skills they deemed necessary for answering the question. Subsequently, we introduced a prompt that instructed the LLM to refine the language of this list, aligning it with the verbs typically used in Bloom's Revised Taxonomy [19]. This step was deliberately conducted after the initial list creation to avoid predisposing the selection of certain verbs, which we found reduced the quality of the labels during pilot testing. For instance, without this step, the experts would typically always suggest skills around the words of "understand", "apply", and "analyze" as they tried to strictly adhere to these levels of Bloom's Revised Taxonomy. Finally, leveraging the insights from the discussion, the refined list of skills and knowledge aligned with Bloom's Taxonomy, and the MCQ itself, the last prompt required the LLM to select the skills or knowledge most pertinent to the question at hand.

Our second prompting strategy, *simulated textbook*, drew inspiration from recent advancements in the creation of knowledge graphs, which frequently employ digital textbooks as a foundational source [10,47]. These methods leverage structural elements like text headers or textbook indexes to outline the initial framework of the graph. Mirroring the expert approach, this strategy contextualizes the task around a MCQ as it might appear in a textbook. The LLM was tasked with identifying the specific, detailed topics a textbook page would cover if it included the given MCQ. Following this, a subsequent prompt directed the LLM to refine these topics, utilizing the language and verb categories found in Bloom's Revised Taxonomy [19]. The final step involved selecting the topic most relevant to the MCQ, ensuring that the identified knowledge areas were directly applicable to the question's context.

---

[2] https://pslcdatashop.web.cmu.edu/DatasetInfo?datasetId=5843
[3] https://github.com/StevenJamesMoore/LearningAtScale24
[4] https://platform.openai.com/docs/models/gpt-4-and-gpt-4-turbo

### 3.3 Human Evaluation

To benchmark this work, we initially compared the original KC tags, which were manually generated by the course creators, to those generated by the LLM. This comparison provided a baseline matching metric for each of the two prompting strategies across both domains. Beyond this direct match metric, our analysis extended to evaluating the outcomes of the second prompt within each prompting strategy. For both prompting strategies, this second prompt generates a list of the *top five* potential KCs for a MCQ, as identified by the LLM. This *top five* list was created with the intention of presenting it to an expert for selection, as part of potential future work. We considered it a partial success when the original manually generated KC tag appeared within this *top five* list, for either of the strategies. This occurrence was documented as a secondary, albeit less precise, metric of matching accuracy.

Recognizing that manually created KCMs can have issues, such as incorrect labeling or wording that misrepresents the required knowledge or skills, we implemented a secondary human evaluation. This aimed to assess the preference between LLM-generated KCs and the original human-generated KCs, specifically for instances of mismatch. This secondary evaluation was only done for the mismatches from the *simulated textbook* strategy, as that had the highest matching percent to the original KC labels across both domains. For example, in a Chemistry MCQ, the human-generated KC was labeled "Use Gay Lussac's law" whereas the LLM-generated KC was "Understand gas pressure-temperature relationship". Given this discrepancy, we asked multiple human evaluators to choose which KC they believed more accurately matched the MCQ.

For this evaluation, we enlisted three domain experts in Chemistry and three in E-Learning. In the case of Chemistry, experts holding a bachelor's degree in Chemistry from the United States were recruited through Prolific, an online research platform [33]. These experts were given a survey implemented via Google Forms that tasked them with reviewing a series of MCQs accompanied by two KC labels. They were instructed to select the label that best matched a set of predefined KC criteria aligned with the KLI framework [17]. These criteria emphasized clarity, direct relevance to the subject matter, factual accuracy, and the ability to apply or integrate the knowledge into broader contexts or practical situations. Specifically, for the 35 MCQs identified as mismatches from the *simulated textbook* strategy in Chemistry, experts evaluated which of the two labels best met these criteria in relation to the MCQ. On average the task took roughly 28 minutes and participants received a compensation of $10.

The same procedure was applied to the E-Learning dataset, adhering to identical instructions and task formats as used for Chemistry. However, the three domain experts involved in this evaluation were instructional staff who had both participated in the E-Learning course and contributed to its development yet had not participated in the original creation of the KCM. For the E-Learning dataset, there were 52 MCQs identified as mismatches based on the *simulated textbook* strategy. On average the task took roughly 32 minutes and participants received a compensation of $15.

An example of a Chemistry and E-Learning question used in both tasks can be seen in Figure 3. To ensure objectivity, the sequence of questions and the presentation order of the two labels for each question were randomized for each expert. To address the issue of LLM-generated KC labels being more verbose than their human-generated counterparts, we utilized the `gpt-4-0125-preview` API to refine these labels for clarity and brevity. We issued a prompt directing the LLM to rephrase each KC, ensuring the revised length did not exceed 1.5 times the word count of the corresponding human-generated KC. For example, a human-generated KC consisting of ten words would result in an LLM-revised KC limited to a maximum of fifteen words. This approach was implemented to equalize the articulation level across labels and align with the typically concise format of KC labels. The preferred label was determined based on a majority vote, where at least two out of three experts had to agree on the choice. This methodological approach aimed to rigorously assess the comparative quality and applicability of LLM-generated KCs against those originally crafted by humans.

> What is the chemical formula for magnesium bromide? *
>
> A) MgBr2
> B) Mg2Br
> C) MgBr
> D) Mg2Br2
>
> ○ Explain balancing charges in ionic compounds, using magnesium bromide
> ○ Formulate binary ionic compound's chemical formulas
>
> Let's identify key features of a good experiment. True or false? A good *
> experimental comparison involves two versions of instruction that vary on only
> one variable.
>
> A) TRUE
> B) FALSE
>
> ○ Define features of a successful experiment
> ○ Emphasize the importance of changing only a single variable in an experiment

**Figure 3: A Chemistry MCQ (top) and E-Learning MCQ (bottom) used in this study.**

### 3.4 Generating KC Ontologies

The prompting strategies described above can be employed to generate KCs for each individual question. One limitation of this methodology is that two questions that assess the same KC can receive LLM-generated labels that both contain the correct semantic information, but that feature two different wordings, as seen in Figure 4. Thus, the resulting KCM might contain redundancies which need to be resolved before using the KCM for purposes of assessing students' KC mastery or problem sequencing.

Which of the following equations should be used to determine the pressure?
A) $P_1V_1 = P_2V_2$
B) $(V_1 / T_1) = (V_2 / T_2)$
C) $(V_1 / n_1) = (V_2 / n_2)$
D) $(P_1 / T_1) = (P_2 / T_2)$

**Human KC**: Apply Boyle's law
**LLM KC**: Examine Boyle's Law

A sample of argon gas is collected in a cylinder with a movable piston as seen in the diagram below. The initial measurements of the gas are given. The piston moves upward and the volume expands to 520.0 mL. What is the new pressure of the sample of gas, assuming that the temperature remains constant?
A) 1.02 atm
B) 2.26 atm
C) 0.0409 atm
D) 25.6 atm

**Human KC**: Apply Boyle's law
**LLM KC**: Utilize Boyle's Law

**Figure 4: Two Chemistry MCQs targeting the same KC, with slight wording variations by the LLM.**

---

**Algorithm 1** KC Ontology Induction

**Given:** question set $Q = \{q_i | i = 1, 2, \cdots, n\}$
**Initialize:** Grouping $G_1 = \{g_1 = Q\}$
1: **for** $t = 1, 2, \ldots$ **do**
2:     Initialize $G_{t+1} = \{\}$
3:     **for** $g_i \in G_t$ **do**
4:         Determine $K_{g_i} = \texttt{determine\_kcs}(g_i)$
5:         Update $G_{t+1} = G_{t+1} \cup \texttt{partition}(g_i, K_{g_i})$
6:     **end for**
7:     **if** $G_t$ equals $G_{t+1}$ **then**
8:         return $G_t$
9:     **end if**
10: **end for**

---

To promote alignment between the generated KCs, we propose an algorithm that induces an ontology of KCs of increasing granularity by iteratively partitioning the question pool into multiple groups. The overall algorithm is presented in Algorithm 1. The algorithm employs two prompts–shown in Figure 5– fulfilling two distinct tasks: (i) Determine a set of learning objectives that can be used to partition the question pool; (ii) Assign each question to one of the learning objectives to form groupings. After grouping the questions, the algorithm uses recursion and continues partitioning the individual subgroups until the LLM learning objectives are of finer granularity or until a group only contains a single question. We phrase the task of identifying KCs as determining fine grained learning objectives to be more aligned with common language. While it seems tempting to directly employ the partitioning induced by the first prompt, we noticed that when working with large questions list (e.g., >50)

GPT-4 can fail to execute the instructions correctly either omitting questions in the assignment process or by assigning the same question to multiple groups. Having an explicit classification prompt that assigns each question to the most relevant group resolves this issue.

```
DETERMINE KCs PROMPT """Below there is a list of
questions and answers intended for a {context}
audience with existing prior knowledge on the subject
of {subject}. You are an educator who sorts the
questions based on learning objectives into groups.
Ensure that each question belongs to EXACTLY one group
(not more or less).

Question List:
{question_list}

Use the following output format:
Group 1 name: [learning objective]
Group 1 questions: [Q1_1, ..., Q1_j]
...
Group N name: [learning objective]
Group N questions: [QN_1, Q_N_k]"""

CLASSIFY QUESTION PROMPT """Below there is a question,
its answer, and a list of learning objectives. You are
a {subject} educator that determines the learning
objective that is most relevant to the question.

Question: {question}

Learning Objectives:
{objectives}

Use the following output format:
Most relevant Objective: [OBJECTIVE NUMBER]"""
```

**Figure 5: Two prompts are used to determine appropriate learning objectives and then to classify the individual questions.**

The iterative partitioning of questions induces an ontology of KCs of increasing levels of granularity. The expert can then decide which level of granularity is most suitable for their application. In this work, we employ the expert labeled datasets to evaluate the quality of our groups. Each dataset features a set of KCs denoted $K$ and a set of questions denoted as $Q$. By design each $k \in K$ is associated with two questions $q_{k,1}, q_{k,2} \in Q$. A question grouping $G$ is characterized by a set of disjoint groups $\{g_1, \ldots, g_n\}$ each hosting a subset of $Q$ (i.e., $g_i \subseteq Q$). The optimal grouping $G^*$ is characterized by $|G^*| = |K|$ and for all $k \in K$ there is an $i \in 1, \ldots, |K|$, such that $g_i = \{q_{k,1}, q_{k,2}\}$. Each step of our algorithm induces a grouping $G_t$. To assess the quality of these groupings at each step we define *grouping accuracy* and *grouping refinement* measures as follows:

$$\mathrm{acc}(G) = \frac{1}{|K|} \sum_{k \in K} \mathbb{1}\left[\exists g_i \in G : q_{k,1} \in g_i \wedge q_{k,2} \in g_i\right] \quad (1)$$

$$\mathrm{ref}(G) = \frac{1}{|Q|} \sum_{g_i \in G} \frac{|g_i|}{|\{k : \exists q \in g_i, \text{s.t. } KC(q) = k\}|} \quad (2)$$

*Grouping accuracy* describes the proportion of question pairs which are correctly co-located in one of the groups. *Grouping refinement* evaluates the average number of KCs in each group. The optimal grouping $G^*$ has an accuracy of 1 and a refinement of 1. The initial dataset that hosts all questions in a single set has an accuracy 1 and refinement $1 / |K|$. We want our algorithm to increase the refinement of the groupings while maintaining a high level of accuracy (i.e., we do not want to split up question pairs of the same KC).

Table 1 displays the token counts and costs associated with two KC generation prompting techniques, expert and textbook, as well as the generation of KC ontologies for each domain.

**Table 1: Summary of token distribution and associated costs for the different prompting approaches.**

| Method | MCQ count | Total Tokens | Prompt Tokens | Completion Tokens | Cost (USD) |
|---|---|---|---|---|---|
| expert | 160 | 462,880 | 307,680 | 155,200 | 8.00 |
| textbook | 160 | 436,480 | 239,040 | 193,120 | 8.00 |
| chemistry ontology | 80 | 104,548 | 97,736 | 6,812 | 3.34 |
| e-learning ontology | 80 | 87,742 | 81,114 | 6,628 | 2.83 |

## 4 RESULTS

In our study, we initially evaluate the effectiveness of the two prompting strategies within the domains of Chemistry and E-Learning. This evaluation is based on how well each strategy's outcomes align with the expert KCM. Subsequently, we explore the preferences of domain experts for the KC labels when discrepancies arise, determining whether they favor labels generated by human experts or those produced by the LLM. Lastly, we examine the performance of our ontology induction algorithm in both domains, focusing on its capability to categorize unlabeled questions by identifying shared KCs.

### 4.1 KCM Match Success

For the first part of this study, we evaluated how well the KCs generated by the LLM aligned with the KCs originally assigned to MCQs by their authors, across two distinct prompting strategies. Our assessment included a direct comparison of the LLM-generated KC to the author-assigned KC for each MCQ. Additionally, we examined the top five KCs proposed by the LLM from the second prompt in both strategies to determine if any of these suggestions matched the author-assigned KC. Furthermore, we explored whether each MCQ was correctly categorized by only one strategy or if both strategies successfully identified the correct KC. The outcomes of these comparative analyses are presented in Table 2.

Our two-proportion z-test comparing the KC match rates of the *simulated textbook* strategy for Chemistry (42/80, 52%) and E-Learning (28/80, 35%) questions revealed a significant difference (*Z=2.698, p=.007*). This indicates a statistically significant better performance of the *simulated textbook* strategy for Chemistry over E-Learning at p < .05, rejecting the null hypothesis of no difference in KC match rates.

**Table 2: For each domain (Chemistry & E-Learning) and strategy (Expert & Textbook), the performance of LLM-generated KCs in relation to the existing KCM. The frequency of direct matches with the human tagged KC; instances where the KC was present in the top five LLM-generated KCs; occasions where a KC was uniquely identified by only one strategy; and cases where both strategies matched the human tagged KC.**

| | Chemistry | | E-Learning | |
|---|---|---|---|---|
| Metric | Expert | Textbook | Expert | Textbook |
| Direct Match | 42/80 (52%) | 45/80 (56%) | 28/80 (35%) | 28/80 (35%) |
| Top Five | 64/80 (80%) | 63/80 (79%) | 45/80 (56%) | 50/80 (63%) |
| Matched Exclusively | 9/80 (11%) | 12/80 (15%) | 9/80 (11%) | 9/80 (11%) |
| Matched by Both | 33/80 (41%) | | 19/80 (24%) | |

We further explored the effectiveness of the *simulated textbook* strategy for identifying KCs across the MCQs in both domains. This analysis focused on the 40 KCs in each domain, where each KC was linked to 2 MCQs, to assess the accuracy of their tagging. In the Chemistry domain, the *simulated textbook* strategy successfully matched both MCQs to their correct KCs in 15 out of 80 cases (19%), correctly matched just one of the two MCQs also in 15 out of 80 cases (19%), and failed to match the KC in either MCQ for 10 out of 80 cases (13%). Similarly, in the E-Learning domain, both MCQs were accurately matched with their KC in 7 out of 80 cases (9%), only one of the two MCQs was correctly matched in 14 out of 80 cases (18%), and both MCQs failed to be matched in 19 out of 80 cases (24%). These results reveal the LLM's variable success rate in precisely identifying KCs through MCQs across different educational domains, suggesting superior performance for Chemistry compared to E-Learning. However, a chi-square test of independence revealed no significant association between the domain and the three aforementioned matching outcome categories ($X^2(2, N=160) = 5.737, p=.057$).

### 4.2 Human KC Preference

For the MCQs in both the Chemistry and E-Learning domains that did not have a successful match with their KC using the *simulated*

*textbook* strategy, we established the preferred KC label through the consensus of three domain expert human evaluators. A KC label was considered preferred only if at least two out of the three evaluators agreed on its selection. From the *simulated textbook* strategy, Chemistry had 35 MCQs where the human and LLM KCs were mismatched, and E-Learning had 52 MCQs.

Within the Chemistry domain, analysis of 35 MCQs revealed a clear preference for the LLM-generated KC labels, which were chosen in 23 out of 35 cases (66%), compared to human-generated labels preferred in 12 instances (34%). Additionally, we observed a substantial level of agreement among the experts, with two-thirds majority agreement (at least two evaluators in agreement) occurring in 25 out of 35 cases (71%), while unanimous agreement (all three evaluators in agreement) was found in 10 cases (29%). Similarly, in the E-Learning domain, upon examining 52 MCQs, LLM-generated labels were preferred in 32 cases (62%), with human-generated labels being chosen in 20 cases (38%). The evaluators demonstrated a clear consensus, with two-thirds majority agreement present in 34 out of 52 instances (65%) and unanimous agreement observed in 18 instances (35%). A comparison of the preferences by domain can be seen in Figure 6. Aggregating preference data from both domains we can verify a statistically significant preference for the LLM-generated KCs. A two-sided binomial test was conducted to assess whether the human evaluators exhibit a preference towards expert or LLM-generated KC labels. For 57 out of 87 evaluated MCQs, the evaluators favored the LLM-generated labels, indicating a statistically significant preference (*p=0.017*).

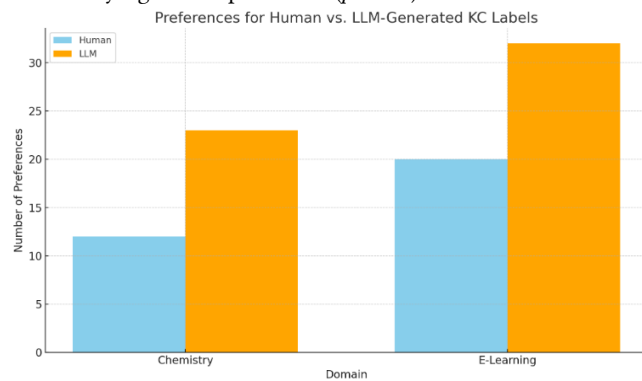

**Figure 6: Comparison of domain expert preferences for human- vs. LLM-generated KC labels.**

### 4.3 Generated KC Ontologies

We now focus on the KC ontologies generated by the clustering algorithm for the Chemistry and the E-Learning datasets. An excerpt of the KC ontology for Chemistry is shown in Figure 7. Going from the root of the tree downwards we can observe how the KCs identified by the algorithm increase in granularity at each step.

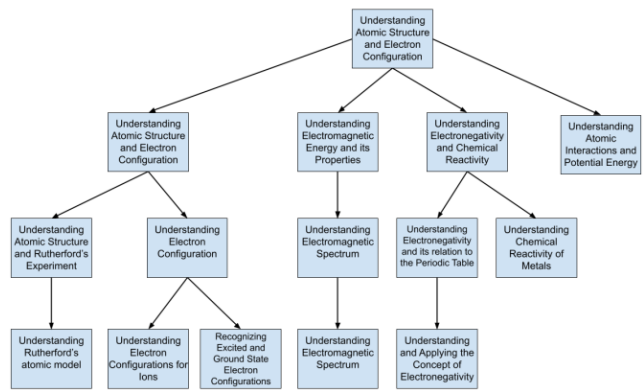

**Figure 7: A section of the tree structure demonstrating KC ontology refinement for part of Chemistry.**

To evaluate the quality of the KCM at different steps we employ the grouping *accuracy* and *refinement* metrics defined in Section 3.4. First, for Chemistry (Figure 8, left) the algorithm converges within 6 iterations to a KCM that groups the 80 questions into 42 different KCs–close to the expert model with 40 KCs. At time of convergence the grouping accuracy indicates that 65% of the question pairs are matched correctly and the grouping refinement of 0.804 indicates that the majority of nodes only feature questions belonging to a single KC. Second, for E-Learning (Figure 8, right) the algorithm converges within 5 iterations to a KCM that groups the 80 questions into 63 different KCs-exceeding the expert model with 40 KCs. At the time of convergence, the grouping accuracy is 17.5% and the grouping refinement is 0.848. Because the final E-Learning KCM employs 63 KCs, many KCs are only tagged to a single question leading to splits between the expert defined question pairs explaining the low accuracy. This suggests that LLM-generated KCM is of finer granularity than the human expert KCM. In real world applications, the domain expert might want to employ a lower level of KC granularity which can be achieved by terminating the algorithm early.

## 5 DISCUSSION

Our results demonstrate the potential of leveraging LLMs to generate and assign high-quality KCs to educational questions. Specifically, in the undergraduate Chemistry domain, we successfully matched more than half of the MCQs with their corresponding KCs, and in the master's E-Learning domain, we achieved a match rate of one-third. For MCQs whose KCs were not directly matched by the LLM, domain expert evaluations showed a two-thirds preference for LLM-generated KCs over the existing human-generated alternatives. Additionally, we introduced a novel clustering algorithm for grouping questions by their KCs in the absence of explicit labels. These results propose a scalable solution for generating and tagging KCs for questions in complex domains without the need for pre-existing labels, student data, or contextual information.

The higher match rate for Chemistry compared to E-Learning can potentially be traced back to the quality of the human-generated

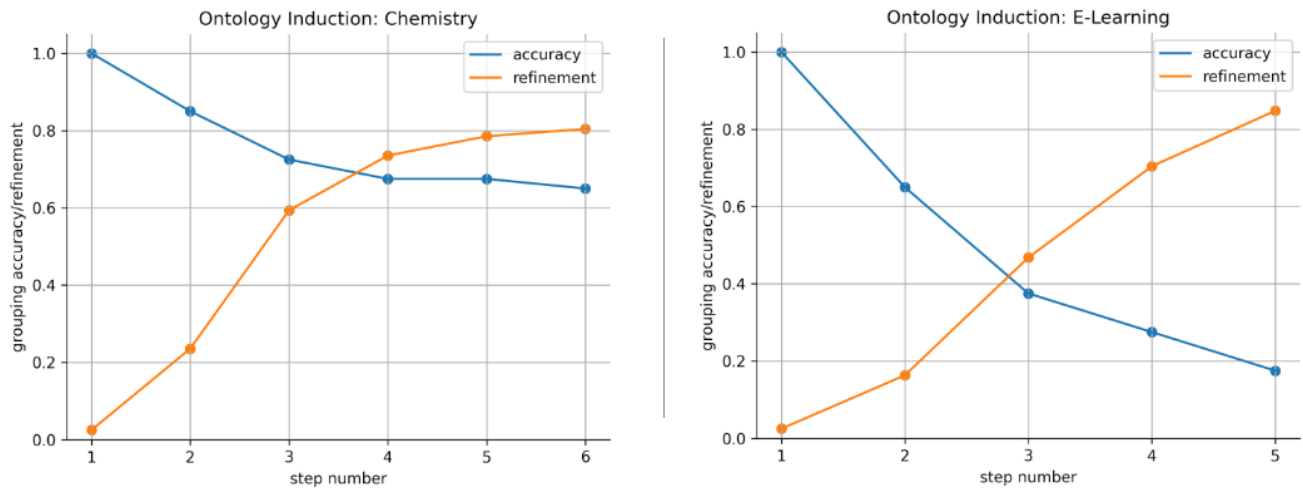

**Figure 8: Grouping quality at different steps of the KC induction algorithm for Chemistry and E-Learning.**

KCM. The Chemistry KCM featured more specific KCs, each typically incorporating just a single piece of domain-specific jargon, unlike the broader KCs with multiple terms found in the E-Learning KCM. Additionally, introductory Chemistry topics are likely to be more prevalent in the LLM's training data than the specialized E-Learning content, which might have contributed to this discrepancy. Despite reasonable success in identifying the top five KCs, the LLM faced difficulties in accurately selecting the most appropriate KC. It often favors general options over precise and domain-specific ones, potentially due to the presence of domain jargon. This led to a substantial portion of MCQs in both domains, 21% in Chemistry and 38% in E-Learning, not matching any of the top five KCs suggested by the LLM. These results indicate that while LLMs are capable of surfacing relevant KCs, the specific nature of domain jargon and a bias towards generalization can impede the accurate identification of a KC.

We observed a notable and statistically significant (two-thirds) preference for LLM-generated KC labels over human-generated ones in both domains. This is possibly due to their slightly longer length and the enhanced readability afforded by the LLM's advanced next-word prediction capabilities. Given this preference, it might suggest the importance of prioritizing human evaluative feedback over direct matches with existing KCMs when assessing LLM effectiveness in generating KCs, especially considering the subjective nature of KC evaluations which can vary significantly based on the reviewer's perspective [22]. Another potential explanation may be that the original KCM designed by the course authors was imperfect, containing KCs that did not fit the particular MCQ or that were too broad. Interestingly, despite the inherent subjectivity in KC evaluation, *all* six reviewers in our study showed a preference for LLM labels. These findings highlight a pronounced preference for LLM-generated KC labels over human-generated ones across both domains. The higher preference for LLM labels, along with the levels of agreement among evaluators, suggests that LLM-generated labels could serve as an effective substitute for manually created labels in categorizing MCQs by their KCs. This preference does not imply that LLM-generated labels should completely replace human input; rather, they could at least provide a valuable foundation, enabling human experts to further select or refine the KCs identified by the LLM. This collaborative approach leverages the strengths of both LLM capabilities and human expertise, potentially leading to more accurate and universally acceptable KC categorizations.

When generating KCs for pairs of questions that assess the same KC, we found that the LLM can assign labels with the correct semantic information, but with different wordings (e.g., see Figure 4). To resolve these redundancies in the KCM, we proposed and evaluated an algorithm that iteratively partitions the question pool to generate KCMs of increasing granularity. The KC ontology induced by this process is similar to taxonomies such as the Common Core State Standards [40] which allow for the categorization of learning materials at different levels of specificity (refer to Figure 7). For the Chemistry questions, we observed that the KCM at the convergence of the algorithm is of similar granularity as the expert model and most expert identified question pairs are grouped correctly. For the E-Learning questions, the converged LLM-generated KCM featured significantly more KCs than the expert model (63 vs 40) indicating a higher granularity. Because of this, our evaluation metrics–which were grounded in the expert KCM–assigned the LLM-generated KCM a low grouping accuracy. Based on the human evaluation of human- and LLM-generated KCs, this might indicate that the expert KCM for the E-Learning course contains inaccuracies and is of lower quality. Lastly, LLM induced KC ontologies might support domain experts structure subject content and provide them with control over the level of KC granularity that is most appropriate. After deciding on a set of KCs the resulting taxonomy could provide a foundation for other types of automated KC tagging algorithms (e.g. [13,34,36]).

Given the preference for LLM-generated KC labels observed in the human evaluations across both domains, practitioners could consider using these labels as initially provided. However, a more effective approach would involve implementing a human-in-the-loop system, where domain experts review and confirm the appropriateness of these labels, creating their own alternatives if necessary. Ideally, this process would start with the LLM-generated labels being preliminarily assigned to problems, followed by a verification step where experts could either approve, modify, or replace them as needed. This process not only ensures accuracy, but also significantly reduces the time and effort required compared to starting from scratch. Ultimately, while the initial LLM-generated labels serve as an effective preliminary pass in developing a knowledge component model, they should be seen as a foundation that can be further refined based on expert insights and student performance data [16].

## 6  LIMITATIONS & FUTURE WORK

In our research, we introduced innovative methods for generating and grouping KCs using a LLM. However, this approach is subject to certain limitations, such as the opaque nature of LLMs, their susceptibility to unexpected output variations, and the potential for biased results [38]. To address these challenges and improve the reliability and efficiency of our methods, we employed a specific iteration of GPT-4, accessed through the `gpt-4-0125-preview` API. This strategy was designed to standardize the evaluation process and guarantee the reproducibility of results by producing consistent outputs in response to predefined prompts. Despite these efforts, the choice of wording in prompts remains a critical factor, significantly affecting the model's output due to LLMs' inherent sensitivity to input nuances. Moreover, the process of evaluating KC quality is complicated by human subjectivity, even among domain experts following specific and detailed guidelines. The definition of a "good" KC is still not clear-cut [17], as reflected in prior literature discussing the desired granularity, making the evaluation process heavily dependent on individual judgment. Our study's scope was limited to two domains, restricted by the scarcity of suitable datasets. Our attempts to use datasets similar to those in prior studies were obstructed due to their unavailability for access. Additionally, niche domains may exhibit poorer performance and higher inaccuracies due to their limited representation in the LLM's training data. For example, this limited representation could explain the low agreement between the expert-created KCM and the E-Learning KC ontology generated by the algorithm.

In future research, we aim to broaden the application of our methods to questions from additional domains and various formats, such as short-answer questions. We are interested in investigating the impact of different contexts, such as instructional text provided before a question, on the quality of KCs generated by LLMs. Our goal is to further refine the prompting strategies we have developed and to foster collaboration among researchers and educators by making our data and code publicly available. Additionally, we plan to explore the potential benefits of utilizing different LLMs, which may enhance the results or provide greater consistency, a notable challenge in current LLM education research.

## 7  CONCLUSION

KCs are crucial for modeling student learning and empowering educational technology with adaptivity and analytics. Therefore, simplifying and scaling the creation and association of KCs with educational content across various domains is essential. In this context, our study suggests that LLMs can play a significant role in facilitating this process. We developed a method to generate KCs for assessment items relying solely on the questions' context and demonstrated its success with assessments from Chemistry and E-Learning courses. Our findings indicate that, although the direct matches between LLM-generated and human-generated KCs were moderate, domain experts most frequently preferred the LLM-generated KCs for the assessments. To overcome the challenge of categorizing assessments by their underlying KCs without labels or context, we also introduced an algorithm for inducing KC ontology and clustering assessments accordingly. Despite the subjectivity, time, and domain expertise that is typically part of the KC mapping process, our approach represents a step towards a scalable solution that addresses these challenges across complex domains. Our research highlights the potential of LLMs to enable individuals, regardless of their technical skills or domain knowledge, to contribute to the development of Knowledge Component Models.

## ACKNOWLEDGMENTS

We thank Microsoft for support through a grant from their Accelerate Foundation Model Academic Research Program. The work of RS and TM was supported by the AFOSR under award FA95501710218.